\documentclass[10pt,twocolumn,letterpaper]{article}

\usepackage{isba}
\usepackage{times}
\usepackage{epsfig}
\usepackage{graphicx}
\usepackage{amsmath}
\usepackage{hyperref}

\usepackage[table]{xcolor}

\isbafinalcopy % *** Uncomment this line for the final submission

\ifisbafinal\fi
\begin{document}

%%%%%%%%% TITLE
\title{  GHCLNet: A Generalized Hierarchically tuned Contact Lens detection Network}

\author{Avantika Singh\\
Indian Institute of Technology Mandi\\
Mandi, India\\
{\tt\small d16027@students.iitmandi.ac.in}
\and
Vishesh Mistry\\
Indian Institute of Technology Jodhpur\\
Jodhpur, India\\
{\tt\small mistry.1@iitj.ac.in }
\and
Dhananjay Yadav\\
Jalpaiguri Government Engineering College\\
Jalpaiguri, India\\
{\tt\small daw123y@gmail.com }
\and
Aditya Nigam\\
Indian Institute of Technology Mandi\\
Mandi, India\\
{\tt\small aditya@iitmandi.ac.in}}

\maketitle
\thispagestyle{empty}

%%%%%%%%% ABSTRACT
\begin{abstract}
Iris serves as one of the best biometric modality owing to its complex, unique and stable structure. However, it can still be spoofed using fabricated eyeballs and contact lens. Accurate identification of contact lens  is must for reliable performance of any  biometric authentication system based on this modality. In this paper, we present a novel approach for detecting contact lens using a Generalized Hierarchically tuned Contact Lens detection Network (GHCLNet) . We have proposed hierarchical  architecture for three class oculus classification namely: no lens, soft lens and cosmetic lens. Our network architecture is inspired by ResNet-50 model. This network works on raw input iris images without any pre-processing and segmentation requirement and this is one of its  prodigious strength.  We  have performed extensive experimentation  on two publicly available data-sets namely: 1)IIIT-D 2)ND and on IIT-K data-set (not publicly available) to ensure the generalizability of our network. The proposed architecture results are quite promising and outperforms the available state-of-the-art lens detection algorithms.

\end{abstract}

%%%%%%%%% BODY TEXT
\section{Introduction}
Security is an important issue for every individual, organization and country to protect it’s information from unauthorized access. Today, a major portion of all information is stored in the form of digital documents. Password based security  has become futile because of its drawbacks such as short passwords which can be cracked easily and the strong ones which are cumbersome to remember, yielding them ineffective. In such situations biometric based authentication system which provide a unique personal identification to all, proves to be a more reliable security system. There are many different traits in the field of biometrics, the selection of a trait is an essential task to increase the robustness of security. Iris is considered as one of the best traits for biometric authentication because it’s complex patterns are unique, stable, and can easily be captured even from short distances, but it can still be spoofed using contact lenses.

The important feature of an iris is it’s textual patterns  which differs from person to person. Studies have proved that these patterns are different even among zygotic twins. The use of contact lens, however, can change these textual patterns. The use of lenses decreases the accuracy of iris detection because of the change in texture brought by them. Therefore, it is of foremost importance to detect the presence of lenses before proceeding for actual iris recognition. Cosmetic lenses, being coloured and textured, are easily detectable as they differ a lot from the texture of a normal iris. However, soft lenses, being transparent in nature, are very difficult to differentiate from no-lens. A lot of  work has been done in the past for lens detection but we are still far behind in accurately differentiating soft-lens from no-lens.

\textbf{Related Work :} The first iris based biometric algorithm was pioneered by Daugman\cite{1} in the late 90's. He suggested a frequency spectrum analysis method to distinguish between a real iris image and a fabricated iris image and to distinguish between an iris without lens and with contact lens.  Zang et al.\cite{2} proposed a method based on weighted Local Binary Patters (LBP) encoded with SIFT descriptors for classifying iris images into lens and no-lens category. Ring et al.\cite{3} detected the regions of local distortion within the iris to detect contact lenses. For this they analyzed the iris bitcode. Doyel et al.\cite{4} ensembled 14 classifiers together to conduct three class lens detection problem and achieved an accuracy of 97\%. Lovish et al. \cite{5} proposed a method based on Local Phase Quantization (LPQ) and Binary Gabor Patterns (BGP) for detecting cosmetic lens. Lee et al.\cite{6} proposed a hardware based solution to distinguish between a real and fabricated iris image based on purkinje image formation. Daksha et al.\cite{7} investigated the effects of texture lens on iris recognition by using variants of Local Binary Patterns. Recently, Ragvendra et al. \cite{20} proposed ContlensNet which is an architecture based on Deep-Convolutional Neural Network for lens detection.
However, it can be concluded from the work done so far that classifying cosmetic lens from no-lens is a well studied problem that achieves a Correct Classification Rate (CCR\%\ ) accuracy upto 99\%\ and above. But accurately differentiating soft lens from no-lens is still a challenging issue.

\textbf{Contribution :} Here in this paper we have used a Generalized Hierarchically tuned Contact Lens detection Network (GHCLNet) for three class ocular classification namely no-lens, soft lens and textured lens. The main contribution of this paper is three fold, that is summarized in the following section.
\begin{enumerate}
	\item Hierarchical Deep Convolutional Network (GHCLNet) for  three class ocular classification namely no-lens, soft lens and cosmetic lens has been proposed. The prodigious strength of this network lies in the fact that  it works on full holistic contact lens features without any pre-processing and  segmentation prerequisite.
	
	\item Generalized deep convolutional neural network based architecture has been proposed.
	
	\item To ensure the generalization ability of the proposed network, multi-sensor and combined-sensor validation has been performed over benchmark databases and compared with the state-of-the-art methods.
\end{enumerate}

 The rest of the paper is organized in the following manner  Section 2 presents the proposed architecture framework, Section 3 discusses the database and testing protocol, Section 4 presents the experimental results and comparative analysis, and Section 5 finally concludes our paper.

\section{Proposed Network}

\begin{figure*}[htp]
	\begin{center}
	
		\includegraphics[width=0.85\linewidth]{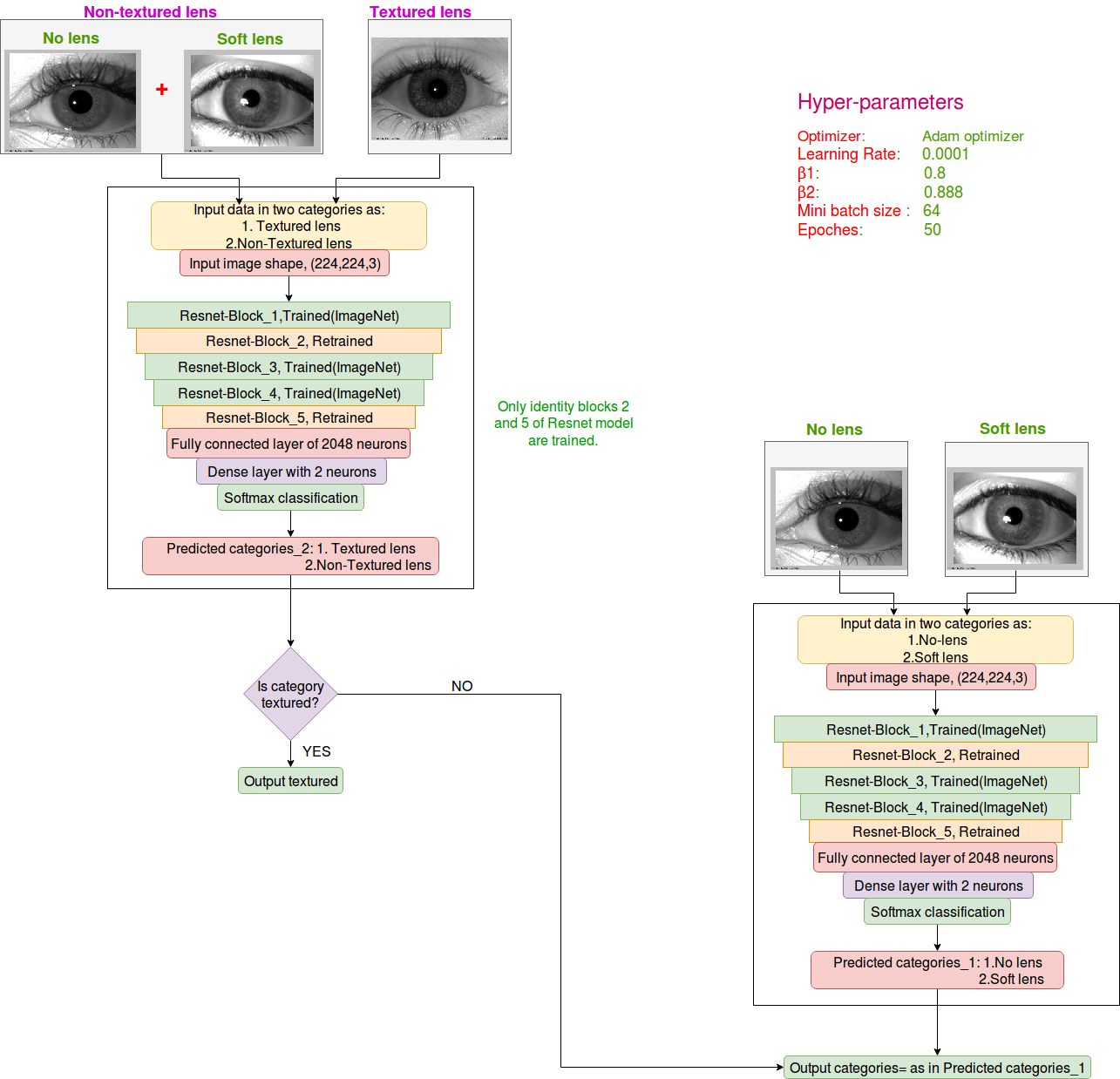}
	\end{center}
	\caption{ Generalized Hierarchically tuned  Contact Lens detection Network (GHCLNet) architecture }
	\label{fig:12}
	
\end{figure*}

Our proposed network is a hierarchical network as shown in Fig.\ref{fig:12} which is inspired from ResNet-50 architecture. The first part of the hierarchical network was exclusively trained for classifying iris images into ‘textured’ and ‘non-textured’ and the second part was exclusively trained for classifying iris images into ‘lens’ and ‘no-lens’. Both the parts were pre-trained ResNet-50 models on ImageNet images with the first part re-trained on textured-lens (‘textured’) images and no-lens + soft-lens (‘non-textured’) images and the second part re-trained on soft-lens (‘lens’) images and no-lens (‘no- lens’) images.
The ResNet-50 model is a popular deep convolutional neural network model made up of five blocks as explained below:
 \begin{itemize}
\item Block-1 is the initial branch which gets the input RGB image of size $224 * 224$. The input image is convolved with $64$ kernels to give a feature map of $112 * 112$, which is then passed to the max-pooling layer
to reduce its size to $55 * 55$.

\item Block-2 comprises of three sub-blocks : $block-2a$, $block-2b$ and $block-3c$. The output feature map of block-2 is of size $55 * 55$.

\item Block-3 comprises of four sub-blocks : $block-3a$, $block-3b$, $block-3c$ and $block-3d$. The final output of block-3 is a $28 * 28$ feature map.

\item Block-4 consists of six sub-blocks namely : $block-4a$, $block-4b$, $block-4c$, $block-4d$, $block-4e$ and $block-4f$. In the end the output feature map is of the size $14 * 14$.
\item  Block-5 is the last block which consists of three sub-blocks : $block-5a$, $block-5b$ and $block-5c$. The output of block-5 is a feature map of size $7 * 7$.
\end{itemize}

\textbf{[a] ResNet Pruning : } ResNet-50 is a very deep network with over $170$ layers pre-trained on the ImageNet database. During experimentation, we found that similar performance was achieved while re-training the 3rd and 5th blocks of ResNet-50 instead of retraining all the blocks. Re-training on any of the 1st, 2nd and 4th blocks resulted in a drastic drop in performance. After extensive experimentation it was found that for the four databases viz. IITK, Cogent, Vista and ND2, maximum performance along with minimal training time was achieved when only the 3rd and 5th blocks of the ResNet-50 models were trained. Since the database of ND1 is very small in number as compared to other databases, the training was only restricted to the sub-block 5(c) in order to avoid over-fitting. The other network parameters that were found after conducting extensive experimentation are summarized in table .\ref{table:10}

During testing, the test image is fed into both the models. The output of the first part of the hierarchical network is first checked. If it classifies the image as ‘textured’, the image is assigned the label of textured-lens. However, if it classifies the image as ‘non-textured’, the output of the second part of the network is considered. If it  classifies it as ‘lens’, the image is put in the soft-lens category and if it classifies it as ‘no-lens’, the image is put in the no-lens category.

 \begin{table}[htp]
 	\begin{center}
 		\begin{tabular}{|p{2.2cm}|p{2.8cm}|}
 			\hline
 			Parameter & Value\\
 			\hline\hline
 			Optimizer  & Adam optimizer\\
 			Learning Rate & 0.0001 \\
 			\(\beta 1 \)& $0.8$ \\
 			\(\beta 2 \) & $0.888$ \\
 			Mini batch size & 64\\
 			Epoches & 50\\
 			\hline
 		\end{tabular}
   	\end{center}
 	\caption{Summarizing the GHCLNet  Parameters}\label{table:10}
 \end{table}

 \textbf{[b] Network Implementation Details : } The proposed generalized hierarchically tuned contact-lens detection network  has been implemented using python and keras\cite{keras}  library using tensorflow\cite{tensor}  as its backend. All the implementation has been done on Intel(R) Xenon(R) CPU E5-2630 V4 at 2.20 GHz, with 32 GB RAM and NVIDIA 1080 Ti gpu with 8 GB RAM.
\section{Database and Testing Protocol}
In this section, we present the details about the database and the testing protocols used. We have tested our proposed network on IIT-Kanpur contact lens iris database (IIT-K)  and on two publicly available iris databases namely: 
Notre Dame cosmetic contact lens 2013 database (ND) and IIIT-Delhi contact lens iris database (IIIT-D). The detailed description about these databases are presented in the following sub section and in table \ref{table:1}

\subsection{IIT-K} This database consists of a total of $12,828$ iris images corresponding to $50$ subjects captured using Vista Imaging FA2 sensor. Since iris images are available for both the left and right eye there are $100$ unique instances of iris present in this database. All the soft lenses used in this database are manufactured by Johnson \& Johnson \cite{18}  and Bausch \& Lomb \cite{15} and all the cosmetic lenses are manufactured by CIBA Vision \cite{14}, Flamymboyout, Oxycolor and FreshLook. This database is provided with an evaluation protocol that comprises of $80$ subjects for training and remaining  $20$ subjects for testing.

\subsection{ND } This database is conceptually divided into two databases namely : ND-I and ND-II. ND-I comprises of $600$ training set images and $300$ testing set images all captured using IrisGuard AD-100 sensor \cite{17}. ND-II comprises of $3,000$ training set images and $1,200$ testing set images all captured using LG-4000 \cite{16} sensor. CIBA Vision\cite{14}, Johnson \& Johnson \cite{18} and Cooper Vision \cite{19} are the three main suppliers of cosmetic contact lenses in this database. In this work for evaluating the proposed framework we follow the evaluation protocol as recommended for this database \cite{4}.

\subsection{IIIT-D } It consists of a total of $6,570$ iris images corresponding to $101$ subjects captured using two sensors namely: Cogent dual iris sensor (CIS 202) and Vista FA2E single iris sensor. All the soft lenses used in this database are manufactured either by CIBA Vision \cite{14} or by Bausch \& Lomb \cite{15}. This database is provided with an evaluation protocol that comprises of $50$ subjects for training and remaining $51$ subjects for testing.

\begin{table}[htp]
 	\begin{center}
 		\begin{tabular}{|p{1.1cm}|p{0.6cm}|p{0.6cm}|p{1cm}|p{0.75cm}|p{0.7cm}|}
 			\hline
 			database &\multicolumn{2}{l|}{Sensor Model}& Total images & Train Set & Test Set\\\hline\hline
 			IITK\cite{5}& \multicolumn{2}{p{3cm}|}{\raggedright Vista Imaging FA2} & 12,823 &10,258 & 2,565 \\ \hline\hline
   		    IIITD\cite{7}& \multicolumn{2}{p{3cm}|}{\raggedright Cogent dual \\ Vista FA2E} & 6,570 & 3,285 &3,285\\ \hline\hline
   	    	ND\cite{13}& \multicolumn{2}{p{3cm}|}{\raggedright LG4000 \cite{16}\\ IrisGuard-AD100 \cite{17}} & 5,100 &3,600 &1,500\\ \hline
 		\end{tabular}
   	\end{center}
 	\caption{Databases used in the proposed GHCLNet architecture}\label{table:1}
 \end{table}

\section{Experimental Results and Discussion}

Our network was trained and tested quantitatively and the results were presented on four different experiments namely: (a) Intra-sensor validation (b) Inter-sensor validation (c) Multi-sensor validation (d) Combined-sensor validation. The quantitative results are presented using the Correct Classification Rate (CCR\%\ ) and thus, the higher the value, the better is the performance.

\subsection{Intra-sensor Validation }
In this testing strategy, training and testing is done for data captured from a single sensor. The results shown in table\ref{table:3} indicates the performance of our proposed network for different sensors. The proposed network's results were analyzed against three other state-of-the-art algorithms namely Statistically Independent Filters\cite{21}, Deep Image Representation\cite{22} and ContlensNet\cite{20}. The following observations were made after the analysis:
\begin{itemize}
\item The results obtained on IIITD  cogent database with the proposed network show the best performance with a total CCR\%\ of 93.71\%\ . The proposed network shows a total hike of 6\%\ in CCR\%\ as compared to the previous state-of-the-art algorithm of ContlensNet \cite{20}.
\item The results on IIITD vista database show exceptional performance of the proposed network with a total CCR\%\ of 95.49\%\ . The results obtained show a total hike of 8\%\ in CCR\%\ as compared with second best ContlensNet model \cite{20}.
\item The results obtained on ND-I database are some what less than the available state-of the art results. The main reason behind this can be attributed to the lesser number of training images available in this database.
\item The results obtained on ND-II database are comparable to the available state of the art technique.
\item The results obtained on IITK databases are exceptional with a total CCR\%\ of 99.67\%\. The accuracy of all the classes are above 99\%\ in CCR\%\ .This can be attributed to the fact that large number of training samples made the network learn better discriminative representations corresponding to each classes.
\end{itemize}

\begin{table*}[htp]
\begin{center}
 
\begin{tabular}{|r|l|r|l|r|l|r|}
	\hline
	Database & Classification Type & SIF\cite{21}\  & DIR\cite{22}\  & ContlensNet\cite{20}\  & GHCLNet\\ \hline \hline
	IIITD- Cogent& N-N&64.16 & 35.50 & 68.68 &89.86 \\ 
	& S-S& 66.45 & 98.21 & 93.62 &91.26 \\
	& C-C& 100 & 73.00 & 100 & 100 \\\cline{2-6} 
	& \textbf{Aggregate}& 76.87 & 69.05 & 86.73 & \cellcolor{green}93.71 \\ \hline \hline
	IIITD- Vista& N-N& 68.89 & 60.80 & 74.50 & 94.6\\ 
	& S-S& 75.63 & 98.30 & 87.50 & 91.88\\ 
	& C-C& 100 & 55.88 & 100 &100 \\\cline{2-6} 
	& \textbf{Aggregate}& 81.50 & 72.08 & 87.33 &\cellcolor{green}95.49 \\ \hline \hline
	ND-I& N-N&76.50 &84.50 &93.25 &91.67\\ 
	& S-S&84.50 &73.75 &97.50 &87.50\\ 
	& C-C&100 &99.75 &100 &100\\\cline{2-6} 
	& \textbf{Aggregate}&87.00 &86.00 &96.91 &\cellcolor{red}93.05\\ \hline \hline
	ND-II& N-N&79.50 &73.00 &88.00 & 95.24\\ 
	& S-S&62.00 &65.00 &97.00 & 89.74\\ 
	& C-C&100   &97.00 &100 & 99.75 \\\cline{2-6} 
	& \textbf{Aggregate}&80.50 &78.33 &95.00 &\cellcolor{red} 94.91\\ \hline \hline
	IITK& N-N&- & - & - & 99.78\\ 
	& S-S& - & - & - & 99.24\\ 
	& C-C& - & - & - & 100\\\cline{2-6} 
	& \textbf{Aggregate}& - & - & - & \cellcolor{green}99.67\\ \hline

\end{tabular}\\
\end{center}
\caption{Intra-Sensor qualitative performance in CCR(\%) on GHCLNet architecture(where N-N is No lens-No lens,S-S is Soft lens-Soft lens , C-C is Cosmetic lens-Cosmetic lens, SIF is Statistically Independent Features\cite{21} and DIR is Deep Image Representation\cite{22}),Green colour represents significant rise in CCR( \%), Red colour indicates no rise in CCR(\%) }\label{table:3}
\end{table*}

\subsection{Inter-Sensor Validation}
In this testing strategy, quantitative performance of the proposed network is shown on inter-sensor validation. The network is trained on one sensor and testing is done on  another sensor. Here, we perform pairwise comparison of IIITD Vista and IIITD Cogent, and ND I and ND II that will result in four different cases. Table \ref{table:4} shows the quantitative performance of the proposed network against three state-of-the-art algorithms namely: Statistically Independent Filters\cite{21}, Deep Image Representation\cite{22} and ContlensNet\cite{20}. The following are the prominent observations based on this experiments:
\begin{itemize}
\item When training data is from IIITD Vista sensor and testing data is from  IIITD Cogent sensor an accuracy of 82.61\%\ in CCR\%\ is obtained on the proposed network.
\item When training of the proposed network is done on data from IIITD Cogent sensor and testing is done on data from IIITD Vista sensor an accuracy of 92.01\%\ in CCR \%\ is noted.
\item When the training data is chosen from ND-II sensor and testing data is chosen from ND-I sensor an excellent accuracy of 91.51\%\ in CCR\%\ is obtained from the proposed network. The results show an increase in accuracy of 3.5\%\ in CCR \%\ as compared to the previous state-of-the-art algorithm of ContlensNet \cite{20}.
\item When the training data is chosen from ND-I sensor and testing data is chosen from ND-II sensor, the best performance of 90.58\%\ in CCR\%\ is obtained from the proposed network. The results show an hike in accuracy of 0.13 \%\ in CCR \%\ as compared to the previous state-of-the-art algorithm of ContlensNet \cite{20}.
\end{itemize}

\begin{table*}[htp]
 \begin{center}
 \begin{tabular}{|r|l|r|l|r|l|r|l|}
	\hline
	
	Train-database & Test-database &Classification Type & SIF\cite{21} &DIR\cite{22}  & ContlensNet\cite{20}  & GHCLNet\\ \hline \hline
    VISTA & COGENT&N-N &57.67 &48.67 &87.75 & 96.74\\ 
	& &S-S&66.06 &42.25 &87.75 & 65.73\\ 
	& &C-C&100 &38.15 &78.91 & 85.36\\\cline{3-7} 
	& & \textbf{Aggregate}&74.57 &43.08 &84.80 & \cellcolor{red}82.61 \\ \hline \hline 
	COGENT & VISTA& N-N&66.91 &06.00 &96.19 &93.40 \\ 
	& &S-S&56.96 &45.47 &88.23 &83.37\\ 
	& &C-C&97.09 &89.61 &100 &99.25\\\cline{3-7} 
	& &\textbf{Aggregate}&73.65 &45.51 &94.80 & \cellcolor{red}92.01\\ \hline \hline
	ND-II& ND-I & N-N &72.66 &75.00 &68.50 &81.25\\ 
	& &S-&54.00 &65.00 &98.00 &93.27\\ 
	& &C-C&100 &94.00 &97.50 & 100\\\cline{3-7} 
	& &\textbf{Aggregate}&75.33 &78.00 &88.00 & \cellcolor{green}91.51\\ \hline \hline
	ND-I& ND-
	II& N-N&57.64 &80.00 &81.33 & 91.9\\ 
	& &S-S&73.64 &49.00 &90.03 & 81.84\\ 
	& &C-C&94.85 &97.00 &100 & 98.00\\\cline{3-7} 
	& &\textbf{Aggregate}&75.37 &75.33 &90.45 & \cellcolor{green}90.58\\ \hline 
			
\end{tabular}\\
\end{center}

\caption{Inter-Sensor qualitative performance in CCR(\%) on GHCLNet architecture(where N-N is No lens-No lens,S-S is Soft lens-Soft lens , C-C is Cosmetic lens-Cosmetic lens, SIF is Statistically Independent Features\cite{21} and DIR is Deep Image Representation\cite{22}), Green colour represents significant rise in CCR( \%),Red colour indicates no rise in CCR(\%) }\label{table:4}
\end{table*}

 \subsection{Multi-sensor Validation}
 In this testing strategy, data from two or more sensors is combined to form a single database. Here, data from same databases is combined to form two separate databases namely: IIITD-combined and ND-Combined. The training data and testing data are combined separately to maintain modality. Table \ref{table:5} indicates the quantitative performance of the proposed network along with two state-of-the-art methods on the multi-sensor validation. The observations made after experimentation are as follows:
\begin{itemize}
\item Training and testing of ND combined data(both ND-I and ND-II) on the proposed network show an excellent accuracy of 95.57\%\ in CCR\%\ . The results show a hike of 3\%\ from the previous state-of-the-art network of ContlensNet \cite{20}.
\item Training and testing of IIITD combined data(both Vista and Cogent) show an accuracy of 94.82\%\ in CCR\%\ obtained from the proposed network. The results obtained show an improvement of 0.17\%\ in CCR\%\ as compared to the previous state-of the art ContlensNet \cite{20}.
\end{itemize} 

 \begin{table*}[htp]
 \begin{center}
 \begin{tabular}{|r|l|r|l|r|l|r|l|}
	\hline
	Database & Classification Type & DIR\cite{22}  & ContlensNet\cite{20} & GHCLNet\\ \hline \hline
	ND-Combined& N-N&77.40 &95.40 & 91.67\\ 
	& S-S&71.40 &82.40 & 95.04\\ 
	& C-C&99.60 &100 & 100\\\cline{2-5} 
	& \textbf{Aggregate}&82.80 &92.60 &\cellcolor{green} 95.57\\ \hline \hline
	IIITD- Combined& N-N&47.55 &96.56 &91.87\\ 
	& S-S&97.99 &88.90 & 92.85\\ 
	& C-C&61.07 &98.50 &99.73\\\cline{2-5} 
	& \textbf{Aggregate}&69.28 &94.65 &\cellcolor{green} 94.82\\ \hline 
			
\end{tabular}\\
\end{center}

\caption{Multi-Sensor qualitative performance in CCR(\%) on GHCLNet architecture(where N-N is No lens-No lens,S-S is Soft lens-Soft lens , C-C is Cosmetic lens-Cosmetic lens and DIR is Deep Image Representation), Green colour represents significant rise in CCR( \%), Red colour indicates no rise in CCR(\%) }\label{table:5}
\end{table*}

\subsection{Combined-sensor Validation}
In this testing strategy, training data of all the databases are combined to form a single large train database while testing is performed on individual test databases. The combined database constitutes of: 1)IITK, 2)IIITD (Cogent and Vista) and 3)ND (ND-I and ND-II). The result obtained by experiments are recorded in Table \ref{table:6} . The following are the observations from the experiment:
testing the network on IITK , ND-I, ND-II, IIITD Vista, IIITD Cogent results in a very good accuracy of 99.14\%, 92.87\%, 94.93\%,  95.69\% and 95.43\% respectively.
Here, we are not able to perform any comparative analysis because this kind of validation was not done earlier. The main aim of doing this validation is to show that our network is trained quite well on different images acquired from different kinds of sensors. This depicts the great generalization ability of our network.

 \begin{table*}[htp]
 \begin{center}
 \begin{tabular}{|r|l|r|l|r|l|r|l|}
	\hline
	database & Classes & IITK & ND-I &ND-II & VISTA &COGENT & Avg\\ \hline \hline
    All Database used for training &N-N	&99.67	&84.38& 94.52	&94.8	&95.19	&93.71\\ 
    &S-S	&97.86	&94.23& 90.26	&92.28	&91.43	&93.21\\ 
    &C-C	&99.88	&100 &100 	&100	&99.67	&99.91\\\cline{2-8} 
    & \textbf{Aggregate}	&99.14	&92.87 &94.93	&95.69	&95.43	&95.61\\ \hline 
			
\end{tabular}\\
\end{center}

\caption {Combined-sensor qualitative performance in CCR(\%) on GHCLNet architecture(where N-N is No lens-No lens,S-S is Soft lens-Soft lens , C-C is Cosmetic lens-Cosmetic lens) }\label{table:6}
\end{table*}
\subsection{Comparative Analysis}
It can be inferred from table \ref{table:3},table \ref{table:4},table \ref{table:5},table \ref{table:6} that our proposed architecture( GHCLNet) is performing far better in terms of CCR\% as compared to the algorithms of pre deep learning era\cite{21}, \cite{22}. To the best of our knowledge ContlensNet\cite{20} a recent research paper, is the only architecture based on deep convolutional neural network for contact lens detection.

In ContlensNet\cite{20} architecture they have used OSIRIS V4.1, a publicly available segmentation tool for iris segmentation and normalization. This tool has limited performance due to occlusion, illumination and other environmental factors. Thus, one has to segment huge number of iris images manually. ContlensNet architecture takes normalized and segmented iris region in the form of patches of size $32*32*1$ as the training input. Since this architecture is not taking into consideration the scalera region of the eye, hence it is not effected by occlusion due to eyelashes.

\begin{figure*}[!htp]
	\begin{center}
		\includegraphics[width=15cm,
			height= 9cm,
			keepaspectratio,]
		{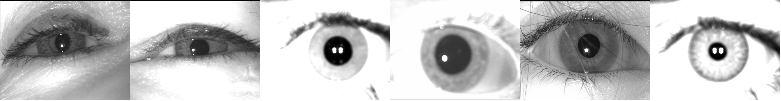}
	\end{center}
	\caption{Input data-set images of poor quality}
	\label{fig:4}
	
\end{figure*}

The main advantage of our proposed GHCLNet is that it is not using any kind of pre-processing and segmentation and still giving comparable results and in many cases even better. Our network is trained in such a way that it is able to handle illumination, occlusion and other external environmental factors in a quite remarkable manner. We are marginally lagging behind ContlensNet at few places as discussed in section 4.1 and 4.2. The main reason for this is the poor quality of raw iris images as can be seen in fig\ref{fig:4}. It is clearly visible from fig\ref{fig:4} that some of the iris images are  illuminated to a large extent which distorts their textual patterns and some of the images are highly occluded. It is very difficult even for a human being to distinguish between no-lens, soft-lens and cosmetic-lens in such kind of images. 
As we are using the entire input raw image in our proposed architecture GHCLNet without segmentation these kinds of factors effect our network performance. But as our network is quite deep when we are combining the data of all data-sets  in consideration we are getting an exceptional high performance as depicted in table \ref{table:6}, this indicates the high generalization ability of our network.

We can summarize our network performance for different testing protocols as follows:
\begin{itemize}
\item\textbf{Intra-Sensor Validation} GHCLNet performance is  quite high in case of IIITD-Cogent and IIITD-Vista but it is less in case of ND-I and ND-II mainly because of less amount of training data available in these datasets and since our network is deep it requires large amount of data for predicting good results.
\item\textbf{Intra-Sensor Validation} It can be observed that GHCLNet performance is quite high from SIF\cite{21} and DIR \cite{22}and marginally lagging behind ContlensNet\cite{20}, that too in few cases mainly because of the poor quality, occluded input images.
\item\textbf{Multi -Sensor Validation} Due to the great generalization ability of our network our results as depicted in table \ref{table:6} outperforms all the available state-of-the art techniques.

\end{itemize}

\subsection{Layer Specific Feature Analysis} 
\begin{figure*}[!htp]
	\begin{center}
		\includegraphics[width=15cm,
			height= 9cm,
			keepaspectratio,]
		{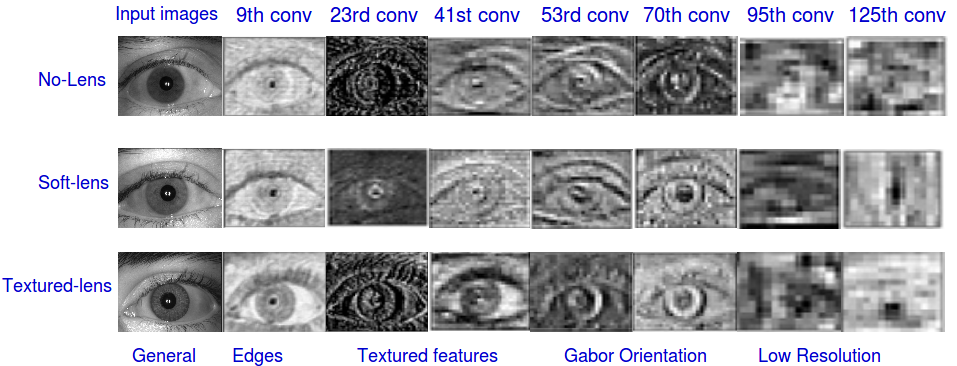}
	\end{center}
	\caption{Features learned by different convolutional layers corresponding to [a] no lens [b] soft lens [c] cosmetic lens}
	\label{fig:3}
	
\end{figure*}

  Fig.\ref{fig:3} shows layer specific feature analysis of no-lens, soft-lens and cosmetic lens images. It is clearly evident from Fig.\ref{fig:3}, that initial convolutional layers learn general specific features. The main reason behind this is that initial convolutional layers look directly at the raw pixels which makes them more interpretative, while as we go deeper features corresponding to no-lens, soft-lens and cosmetic-lens are learned. Features learned by initial layers like $Conv-2$ layer are very basic features, but as we move deeper in the network more specific learning is been performed like in the $Conv-9$ layer which detects edges and lines. $Conv-23$ layer is playing a major role in differentiating no-lens image from cosmetic-lens image. In this layer textual features are learned. Interestingly, we have observed that our network automatically learns state-of-the-art gabor filter like features at different orientations. The lower layers of the network learn high level aggregated discriminative features, as shown in Fig \ref{fig:3}, like $Conv-95$ and $Conv-125$ layers. In the lower layers of the network, the resolution and features become mostly an encoding of few discriminative intrinsic information. 

  \section{Conclusion}
 
 Iris is considered as one of the best traits for biometric authentication as its complex patterns are unique and stable. However, the use of lenses decreases the accuracy of iris detection because of the change in texture brought by them and thus it is required to detect the presence of lenses before proceeding for actual iris recognition. In this paper, we proposed a novel  Generalized Hierarchically tuned Contact Lens detection Network (GHCLNet). 
 
  Extensive experimentation has been carried out with three  publicly available databases using four testing strategies: intra-sensor validation, inter-sensor validation, multi-sensor validation, and combined-sensor validation. The consistent CCR(\%) improvements in multi-sensor validation; and the amazing combined-sensor validation results largely indicates the generalization ability of our network. To the best of our knowledge this kind of combined sensor testing is not done by anyone so far. The proposed architecture, with its promising results, has majorly outperformed the current state-of-the-art techniques. The main strength of this network lies in the fact that it is not using any kind of pre-processing and iris segmentation, and still giving remarkable results. This saves lots of computational  time and can thus  be integrated very easily  as the first step in any iris recognition system to increase its performance.

{\small
\bibliographystyle{ieee}
\bibliography{submission_example}

\begin{thebibliography}{10}\itemsep=-1pt

\bibitem{15}
Bausch \& lomb, rochester, ny, usa. (2014, jan.). bausch \& lomb lenses,
  available: \url{http://www.bausch.com}.

\bibitem{19}
C. vision. (2013, apr.). expressions colors, available:
  \url{http://coopervision.com/contact-lenses/expressions-color-contacts}.

\bibitem{14}
Cibavision, duluth, ga, usa (2013, apr.) freshlook colorblends,
  available:\url{http://www.freshlookcontacts.com}.

\bibitem{17}
Irisguard, washington, dc, usa. (2013, apr.). ad100 camera, available:
  \url{http://www.Irisguard.com/uploads/AD100ProductSheet.pdf}.

\bibitem{18}
Johnson\& johnson, skillman, nj, usa. (2013, apr.). acuvue2 colours, available:
  \url{http://www.acuvue.com/products-acuvue-2-colours}.

\bibitem{16}
Lg, riyadh, saudi arabia. (2011, oct.). lg 4000 camera, available:
  \url{http://www.lgIris.com}.

\bibitem{keras}
F.~Chollet et~al.
\newblock Keras.
\newblock \url{https://github.com/fchollet/keras}, 2015.

\bibitem{13}
J.~Daugman.
\newblock How iris recognition works.
\newblock In {\em Proceedings of the 2002 International Conference on Image
  Processing, {ICIP}}, pages 33--36, 2002.

\bibitem{1}
J.~Daugman.
\newblock Demodulation by complex-valued wavelets for stochastic pattern
  recognition.
\newblock {\em Int. J. Wavelets, Multiresolution Inf. Process.}, 1(1):1--17,
  2003.

\bibitem{4}
J.~Doyle, K.~Bowyer, and P.~Flynn.
\newblock Variation in accuracy of textured contact lens detection based on
  sensor and lens pattern.
\newblock In {\em 6th IEEE International Conference on Biometrics, Technol.,
  Appl., Syst.}, pages 1--7, 2013.

\bibitem{6}
E.~C. Lee, K.~R. Park, and J.~Kim.
\newblock Fake iris detection by using purkinje image.
\newblock In {\em International Conference on Biometrics,IAPR}, pages 397--403,
  2006.

\bibitem{5}
Lovish, A.~Nigam, B.~Kumar, and P.~Gupta.
\newblock Robust contact lens detection using local phase quantization and
  binary gabor pattern.
\newblock In {\em 16th International Conference, Computer Analysis of Images
  and Patterns {CAIP}}, pages 702--714, 2015.

\bibitem{tensor}
A.~Mart et~al.
\newblock Tensorflow: Large-scale machine learning on heterogeneous systems.
\newblock \url{https://www.tensorflow.org/}.

\bibitem{21}
R.~Raghavendra, K.~B. Raja, and C.~Busch.
\newblock Ensemble of statistically independent filters for robust contact lens
  detection in iris images.
\newblock In {\em Proceedings of the 2014 Indian Conference on Computer Vision
  Graphics and Image Processing}, page~24, 2014.

\bibitem{20}
R.~Raghavendra, K.~B. Raja, and C.~Busch.
\newblock Contlensnet: Robust iris contact lens detection using deep
  convolutional neural networks.
\newblock In {\em 2017 {IEEE} Winter Conference on Applications of Computer
  Vision,{WACV} , Santa Rosa, CA, USA}, pages 1160--1167, 2017.

\bibitem{3}
S.~Ring and K.~Bowyer.
\newblock Detection of iris texture distortions by analyzing iris code matching
  results.
\newblock In {\em BTAS}, pages 1--6, 2008.

\bibitem{22}
P.~Silva, E.~Luz, R.~Baeta, H.~Pedrini, A.~X. Falcao, and D.~Menotti.
\newblock An approach to iris contact lens detection based on deep image
  representations.
\newblock In {\em Graphics, Patterns and Images (SIBGRAPI)}, pages 157--164,
  2015.

\bibitem{7}
D.~Yadav, N.~Kohli, J.~S.~D. Jr., R.~Singh, M.~Vatsa, and K.~W. Bowyer.
\newblock Unraveling the effect of textured contact lenses on iris recognition.
\newblock {\em {IEEE} Trans. Information Forensics and Security,2014},
  9(5):851--862, 2014.

\bibitem{2}
H.~Zhang, Z.~Sun, and T.~Tan.
\newblock Contact lens detection based on weighted lbp.
\newblock In {\em 20th International Conference on Pattern Recognition}, pages
  4279--4282, 2010.

\end{thebibliography}
}

\end{document}